\documentclass{article}
\PassOptionsToPackage{numbers,square,sort}{natbib}
\usepackage[final]{neurips_2024}

\usepackage[dvipsnames]{xcolor}
\usepackage[T1]{fontenc}
\usepackage[utf8]{inputenc}
\usepackage{amsfonts}
\usepackage{amsmath}
\usepackage{amssymb}
\usepackage{booktabs}
\usepackage{graphicx}
\usepackage{microtype}
\usepackage{multirow}
\usepackage{nicefrac}
\usepackage{pifont}
\usepackage{soul}
\usepackage{url}
\usepackage{wrapfig}
\usepackage{xfrac}
\usepackage{caption}

\usepackage[pagebackref,breaklinks,colorlinks]{hyperref}
\usepackage[capitalise]{cleveref}

\makeatletter
\newcommand{\cmark}{\ding{51}}%
\newcommand{\xmark}{\ding{55}}%
\makeatother

\newcommand{\ie}{i.e.,\ }
\newcommand{\eg}{e.g.,\ }

\title{Learning segmentation from point trajectories}

\begin{document}
\author{Laurynas Karazija, Iro Laina, Christian Rupprecht, Andrea Vedaldi \\
Visual Geometry Group\\
University of Oxford\\
Oxford, UK\\
{\tt\small \{laurynas,iro,chrisr,vedaldi\}@robots.ox.ac.uk}
}

\maketitle

\begin{abstract}
  We consider the problem of segmenting objects in videos based on their motion and no other forms of supervision.
  Prior work has often approached this problem by using the principle of common fate, namely the fact that the motion of points that belong to the same object is strongly correlated.
  However, most authors have only considered instantaneous motion from optical flow.
  In this work, we present a way to train a segmentation network using long-term point trajectories as a supervisory signal to complement optical flow.
  The key difficulty is that long-term motion, unlike instantaneous motion, is difficult to model -- any parametric approximation is unlikely to capture complex motion patterns over long periods of time.
  We instead draw inspiration from subspace clustering approaches, proposing a loss function that seeks to group the trajectories into low-rank matrices where the motion of object points can be approximately explained as a linear combination of other point tracks.
  Our method outperforms the prior art on motion-based segmentation, which shows the utility of long-term motion and the effectiveness of our formulation.
\end{abstract}

\section{Introduction}\label{sec:intro}

Segmentation, the task of delineating and isolating distinct objects, is a fundamental problem in computer vision. Much of the current approaches are supervised, relying on expensive manual annotations. Attempts to approach this task without supervision have largely relied on manual heuristics or exploited the rich semantics of self-supervised feature extractors. Video data, however, offers an additional option as it contains \emph{motion}, which can be exploited for an additional inductive bias. Such approaches are rooted in the principle of common fate from Gestalt psychology~\citep{wertheimer1912experimentelle}, which posits that elements that move together are more likely to belong together. 

Motion information is usually captured by optical flow. Flow is attractive as it arises from low-level visual properties and can provide a signal before scenes are parsed and objects are discovered. Furthermore, optical flow estimators, such as RAFT~\cite{teed2020raft} or FlowFormer~\cite{huang2022flowformer}, can be trained purely on synthetic artificial data, transferring to real-world scenes with remarkable accuracy and without manual annotation. This has led many to consider optical flow as a critical modality to discover and learn objects from video data by learning to attribute and explain the motions of objects. 

Optical flow, however, only describes the instantaneous motion of the scene, which can create blindspots: not all objects are necessarily in motion at all times. Similarly, groups of objects might coincidentally move together. Recent advances in point tracking~\cite{karaev2023cotracker, doersch2023tapir, doersch2022tapvid, harley2022particle} offer an alternative form of motion information. Point trackers ``lock on'' to a set of query points and describe their position and visibility over the course of the whole video. This provides long-term motion information. Like optical flow estimators, point trackers are trained on synthetic data. However, unlike optical flow, point trajectories describe only a sparse set of points. 

In this paper, we ask whether the long-term motion information obtained in point trajectories is beneficial. To that end, we explore how to supervise image segmentation networks using motion information with point trajectories. At a glance, this presents several problems. Firstly, point trajectories are time-varying 2D point clouds, and combining them with image-based networks is not straightforward. Furthermore, the evolution of long-term object motion is too complex, even in the simplest cases. Our main insight is that the motion of points belonging to the same object should be well correlated. We thus propose a loss function that encodes this intuition by seeking to explain groups of points as combinations of other points in the group. With our method, a segmentation network predicts objects in the scene, inducing a grouping of trajectories that are currently visible. The loss function then assesses how well such a grouping explains the long-term motion. While point trajectories describe motion over a longer time, they are limited by the number of points tracked, which is often much less than the number of pixels. We thus propose to train using both trajectory-based loss and optical flow-based loss and show that spatially sparse but longer-time motion information synergises with spatially dense optical flow.

Discovering objects using point trajectories has a long history in computer vision. Our approach is inspired by ideas of subspace clustering, which assume that data comes from distinct subspaces and seek to reconstruct membership information of data points. This has previously been applied to the problem of motion segmentation \cite{liu2012robust,elhamifar2013sparse}.
These approaches, however, are sensitive to noise and either rely on specialised optimisation procedures to recover a graph of trajectory relationships \cite{ochs2011object,keuper2016multi} or use manual instead \cite{ochs2011object,keuper2016multi}.
Normalised cuts or spectral clustering are then used to group the trajectories.
However, the need for an affinity matrix limits the number of trajectories that can be used due to quadratic memory requirements.
Furthermore, %
``densification'' is still required to extend trajectory clusters to the whole image. 
By construction, these approaches can process only a single sequence at a time.
Our proposal instead trains an image segmentation network directly end-to-end using a dataset of videos while supporting a large number of trajectories. 

In summary, our work makes the following contributions.
(1) We propose a loss function that enables training any image segmentation architecture using point trajectories as a source of supervision. (2) We investigate our proposed loss in a principled way in a simulated setting, showing the feasibility of our approach. (3) We apply such a loss in a per-sequence optimisation, outperforming other subspace clustering baselines.
(4) We use our loss to train a single network on a dataset of videos for the task of video object segmentation, demonstrating strong results. 
(5) We show how our proposed loss formulation obtains better performance than alternatives. 
\section{Related work}\label{sec:related}

\paragraph{Unsupervised video object segmentation.}
Video object segmentation (VOS) aims to label pixels of objects in a video.
Current VOS benchmarks~\cite{perazzi2016a-benchmark-davis,li2013video-segtrack,ochs2014segmentation-fbms} usually define the problem as binary foreground-background separation or salient object segmentation.
The task is usually approached in two ways: semi-supervised and unsupervised VOS. Semi-supervised methods require initial frame annotations and aim to \emph{propagate} them to the rest of the video~\cite{caelles2017one}. Unsupervised VOS aims to discover object(s) of interest without the initial targets~\cite{faktor2014videonlc,papazoglou2013fast,tokmakov2019motion,jain2017fusionseg,li2018instance,lu2019see}. This however does not differentiate methods based on data used to \emph{train} them. Most of the traditional research in semi- or unsupervised VOS relies on annotations during training. Our approach, in contrast, does not rely on any manual annotations to learn. Some authors explore related unsupervised video instance segmentation~\cite{wang2023videocutler} task without any annotations, object-centric learning appraoches~\cite{singh2022simple,zadaianchuk2024object,aydemir2024self}, some of which make use of flow~\cite{karazija2022unsupervised} and depth~\cite{safadoust2023multi}.

\paragraph{Motion segmentation.}
A closely related task to video object segmentation is motion segmentation, which aims to extract the main moving objects in a video. 
The practical difference between these two tasks is more difficult to delineate as the same benchmark datasets are often used. 
Early works modeled the scenes as layers~\cite{chang2013topology,jojic1993learning}, which later works accomplish using a slot-attention mechanism~\cite{yang2021self-supervised,ding2022motion,lao2023divided}.
Flow mixture models accounted for multiple motion patterns~\cite {jepson1993mixture,torr1998geometric}, and corrections were introduced for rotating cameras~\cite{bideau2016s,bideau2018best}. Later works \cite{meunier2022em-driven,meunier2023unsupervised,choudhury+karazija22gwm} considered parametric flow models fit to explain the scene.
AMD~\cite{liu2021emergence} employs a single model with separate appearance and motion `pathways'. 
Other works train flow-only models by generating synthetic data, which generalise well to real videos \cite{Xie2022SegmentingMO, Lamdouar2021SegmentingIM}.
An alternative line of work adopts a more generative approach, training an inpainter networks to predict optical flow~\cite{yang-loquercio2019unsupervised,yang2021dystab}.
Several authors~\cite{lian2023bootstrapping,singh2023locate} adopt a multi-stage self-labelling~\cite{wang2023videocutler} approach for motion segmentation: initial masks are estimated using an optical flow-based approach, followed by DINO-based refinement and CRF post-processing to generate pseudo-labels and train a final segmentation network.

\paragraph{Trajectory-based motion segmentation.}
Trajectory-based motion segmentation has also been explored. Older works consider data of multiple trajectories and employ non-negative matrix factorization and related decomposition methods~\cite{cheriyadat2009non,costeira1995multi,elhamifar2013sparse,fradet2009clustering,rao2008motion,yan2006general}. This line of work primarily operates by defining affinity between pairwise trajectories in a single video setting. In \cite{brox2010object, ochs2012higher, ochs2013segmentation,keuper2015motion,keuper2017higher},  heuristic graphs are constructed between trajectories, considering increasingly complex motion models, and employing specialised solvers to solve the optimisation problem.
However, due to the specialised optimisation procedures and tight coupling with trajectory estimation methods, this line of work has received less attention than deep methods that exploit optical flow similarly to RGB frames. 

\paragraph{Subspace clustering.}
A specific kind of trajectory-based technique is subspace clustering approaches, which rely on the \emph{self-expressive} property of the data. 
They can largely be summarised~\cite{haeffele2020critique} as solving a constrained optimisation problem $\arg \min_C ||DC-D||^2_F + \lambda \theta(C)$ for some dataset $D \in \mathbb{R}^{d \times n}$ of $n$ points in $d$ dimensions. $C$ is a matrix of coefficients, which expresses the data and can be represented as a linear combination of other points. Given a solution for the coefficient matrix, it is transformed into an affinity matrix for spectral clustering. The approaches mainly differ in the second term of the objective and specialized methods to solve the optimisation problem. SSC~\cite{elhamifar2013sparse} define $\theta(C)$ as $l_1$ norm. 
LLR~\cite{liu2012robust} use nuclear norm instead, while LSR~\cite{lu2012robust} uses instead $l_2$ regularisation.
\cite{lu2012robust,luo2011multi,vidal2014low} combines $l_1$, $l_2$, and nuclear norms.
Under some strong assumptions~\cite{haeffele2020critique}, these approaches enjoy some theoretical guarantees.
However, they are difficult to scale in practice as the number of points $n$ grows, as $C$ is $n \times n$. 
Additionally, the secondary step of spectral clustering is also limiting and difficult to tune.
Instead, we take inspiration from these approaches and propose a way to supervise the network directly using the self-expressive property of point trajectories.

\section{Method}\label{sec:method}

Our goal is to solve the video segmentation task in an unsupervised manner: given a video, we want to segment out the objects that are moving independently within it.
A video 
is a sequence of frames
$
\mathcal{I}_t\in \mathbb{R}^{HW \times 3},
$
each of which is an RGB image defined on the lattice 
$
\Omega = \operatorname{vec}(\{1,\dots,H\} \times \{1,\dots,W\})\in \mathbb{R}^{HW \times 1}.
$
To segment the objects, we self-supervise a neural network $\Phi$ that takes as input each frame $\mathcal{I}_t$ in turn, and outputs a corresponding segmentation mask $\Phi(\mathcal{I}_t) = M_t \in [0, 1]^{HW\times K}$ where $K$ is the number of possible segments we expect to observe in the video. 
Segmentation matrix entries softly assign each pixel to one of $K$ possible segments.

The challenge is how to supervise the network $\Phi$ without labels, utilising only the video itself as training material.
The key inductive principle that we propose to use is that physical points that belong to the same object tend to have highly correlated motion, often called \emph{principle of common fate}.
When these points are projected to pixels, they result in corresponding highly correlated apparent motions, which we can measure using techniques like optical flow and point tracking.
Therefore, we propose to supervise the network $\Phi$ from an analysis of apparent motion extracted automatically from the video using off-the-shelf components.

Motion can be measured at two temporal scales.
Optical flow extracts instantaneous motion, measuring the 2D velocity of the 3D points found at each pixel in each video frame.
Point tracking extracts long-term motion, estimating the 2D location of a certain number of 3D points throughout the video's duration.
These two sources of information are complementary.
Optical flow is dense, easy to extract, and easy to model to discover correlations within it; however, by considering different times in isolation, it ignores most of the correlations that exist in the data.
Tracks are sparse, more difficult to extract and harder to model, but potentially contain information ignored by optical flow.

Prior works such as~\cite{choudhury+karazija22gwm} have studied how to model optical flow for segmentation.
Here, motivated by a new generation of high-quality point trackers~\cite{karaev2023cotracker, doersch2023tapir, doersch2022tapvid, harley2022particle}, we aim at developing the machinery necessary to use track information as well.
From this analysis, we construct losses 
which assess the quality of the predicted mask $M_t$ given the video itself.
Next, we introduce two such losses, one for optical flow from prior work, and a new one based on point tracking.

\subsection{Learning from optical flow}

First, we describe the case of optical flow.
Because optical flow is instantaneous, we can fix our attention on a specific frame $\mathcal{I}$ and corresponding mask $M$, dropping for now the time index $t$.
The \emph{optical flow}
$
F \in \mathbb{R}^{HW \times 2}
$
for this image associates a 2-dimensional flow vector to each of the $H\times W$ pixels.
Each flow vector can be understood as the velocity of the pixel.

Let $M_k \in \mathbb{R}^{HW\times 1}$ be the binary matrix for segment $k$, obtained by extracting the $k$-th column of $M$.
Let $F_k = M_k \odot F$ denote the Hadamard (element-wise) product between the mask and flow vectors, broadcasting the mask along the rows.

Assuming that the object is rigid, the optical flow can be approximated as a linear parametric model of 2D coordinate embeddings (see~\cite{adiv1985determining} for an overview).
Following~\cite{choudhury+karazija22gwm}, we consider a six-dimensional quadratic embedding kernel
$
\operatorname{emb}([x,~y])= [x,~x^2,~y,~y^2,~xy,~1] \in \mathbb{R}^{1\times6}
$
for pixel coordinates $[x,~y] \in \Omega$ and associate to each region $k$ a corresponding set of 12 parameters
$
\theta_k \in \mathbb{R}^{6 \times 2}.
$
Optical flow vectors within a region should be expressible as a linear combination of these six basis functions.

We then consider all pixels embeddings stacked in a single matrix
$
E_k = M_k \odot \operatorname{emb}(\Omega)
$
where the product with the soft mask ensures that the embeddings are ``active'' only if the corresponding pixels are.
The optical flow vectors in the region are then approximated as
\begin{equation}
F_k
\approx
\hat{F}_k 
= 
E_k \hat{\theta}_k
\quad\mathrm{where}\quad
\hat{\theta}_k = (E_k^\top E_k)^{-1}E_k^\top F_k,
\end{equation}
where $\hat\theta_k$ is obtained via least square.
We can use the residual of this approximation as a measure of how well the mask $M_k$ fits the data:
\begin{equation}
\label{eq:flow_loss}
\mathcal{L}_f(M|F) 
= \sum_k \| F_k - \hat{F}_k \|^2_F
= \sum_k \| F_k - E_k \hat{\theta}_k \|^2_F\,.
\end{equation}%
Intuitively, this considers the correlation of pixel motion in the \emph{spatial} sense: how pixel coordinates determine its motion based on motion parameters $\theta_k$.

\subsection{Learning from trajectories}

Having covered optical flow, we move now to developing an analogous loss for tracking.
We write $P \in \mathbb{R}^{2T \times N}$ for the track matrix, with one trajectory per column.
With slight abuse of notation, we write $(P)_t \in \mathbb{R}^{2\times N}$ for indexing rows corresponding to point locations at some time $t$.
To connect pixel-wise masks and sparse points, we use a sampling operation $\pi(\cdot)$, writing $\pi(M_k, (P)_t) = \hat{M_k} \in [0,1]^{N\times 1}$ for mask values at point locations at an appropriate time.
Furthermore, we denote by $P_k = P \odot \hat{M}_k$ the masked version of the trajectory matrix, selecting the columns/trajectories that belong to object $k$ with obvious broadcasting of the mask values.

\begin{figure}[t]
    \centering
    \includegraphics[width=0.95\linewidth]{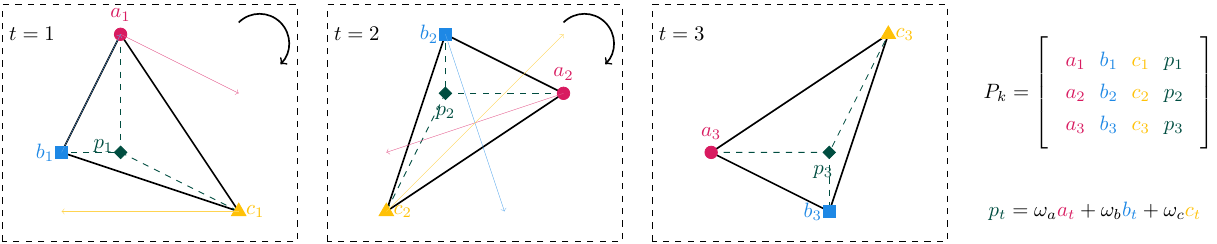}
     \caption{
     \textbf{Illustrative 2D example for the low-rank nature of $P_k$.} 
     A triangle undergoes rigid rotation over three frames. 
     As the rate of rotation is not constant, the flow vectors and point positions are difficult to model. 
     However, the point $p$ is part of the triangle and can be expressed as a combination of the three vertices at an appropriate time. 
     Thus, the last column of $P_k$ is linearly dependent, and $P_k$ is rank deficient. 
     \emph{Any points in the triangle could be included in $P_k$ without increasing its rank.}
     }
    \label{fig:intuition}
\end{figure}

Unlike optical flow, trajectories are too complex to be modelled using a small set of \emph{fixed} basis functions.
Instead, we posit that the set of trajectories should be low-rank\,---\,all trajectories belonging to the same object should be explained well by a linear combination of some small number of trajectories. 
We illustrate this intuition in \cref{fig:intuition} using a 2D example.

This assumption results in a factorization of $P_k$ using singular value decomposition (SVD) as
$
P_k = U_k \Sigma_k V_k^\top
$,
where
$
(U_k,\Sigma_k,V_k) = \operatorname{SVD}(P_k).
$
As $P_k$ should be low-rank, we can thus form an approximation using truncated SVD, by considering only first $r$ components. 
We write $\lfloor U_k\rfloor_r$ to denote such truncation.
With this, we obtain the loss
\begin{equation}\label{eq:loss_rec}
\mathcal{L}_{\mathrm{rec}@r}(M | P)
=
\sum_k 
\left\|
P_k - 
\lfloor U_k\rfloor_r
\lfloor\Sigma_k\rfloor_r
\lfloor V_k\rfloor_r^\top
\right\|^2_F.
\end{equation}
Since truncated SVD offers optimal decomposition for the error above, lowering this loss amounts to making $P_k$ as close as possible to rank $r$, i.e., by grouping trajectories into $P_k$ that do not increase its rank, and should come from rigid objects. 

As we show in \cref{sec:alt_losses}, we found an alternative formulation 
of this idea works better.
Note the rank $r$ matrix has the $r$-th and all later singular values as 0. 
We can optimise singular values higher than $r$-th to be close to 0 (ignoring $U_k$ and $V_k$).
Thus, for trajectories, we formulate a loss simply as:
\begin{equation}\label{eq:traj_loss}
\mathcal{L}_t(M|P) = \sum_k \sum_{i=r}^{ \min (2T, N)} \sigma_i(P_k),
\end{equation}
where $\sigma_i(P_k)$ is the i-th singular value of $P_k$. 
We assume $r \ll \min (2T, N)$.

\paragraph{Meaning of decomposition.} 
We show that under certain simplifying assumptions, the decomposition in \eqref{eq:loss_rec} is exact and models time-varying camera motion and object geometry as two terms. 
We consider a simple case of a rigid body motion observed through a perspective camera. 
For points on the object, we can consider only the relative motion between the camera and the object and attribute it all to the camera for simplicity. 

Given (stacked) camera projection matrices $W_t\in \mathbb{R}^{3T\times 4}$, points $\tilde{X}_k \in \mathbb{R}^{4\times N}$ in homogenous coordinates that remain at constant projective depth $\mathbf{d} \in \mathbb{R}^{N \times 1}$ from the camera over the whole sequence, we note the following equation \cite{Hartley2004}:
\begin{equation}\label{eq:rec_4}
    \tilde{P}_k = W_{t} \tilde{X}_k \operatorname{diag}(\mathbf{d})^{-1} ,
\end{equation}
where $\tilde{P}_k \in \mathbb{R}^{3T \times N}$ is $P_k$ in homogenous coordinates. Both $W_{t}$ and $\tilde{X}_k \operatorname{diag}(\mathbf{d})^{-1}$ can be recovered by considering a truncated SVD at rank 4: $W_t=\lfloor U_k\rfloor_4 \lfloor\Sigma_k\rfloor_4$, and $\tilde{X}_k \operatorname{diag}(\mathbf{d})^{-1}=\lfloor V_k\rfloor_4^\top$.

The trajectory matrix factorises into the time-varying camera matrices and object geometry. As the depth is not constant in the real-world setting, this decomposition is approximate and suggests the following alternative loss:
\begin{align}
\mathcal{L}_\mathrm{per} &= \sum_k \| \tilde{P_k} - W_{t} \tilde{X}_k \operatorname{diag}(\mathbf{d})^{-1}\|^2_F, \label{eq:loss_per} 
\end{align}
where $W_t$, and $\tilde{X}_k \operatorname{diag}(\mathbf{d})^{-1}$ are obtained via SVD as above.

\paragraph{Choice of $r$.} 
Setting $r$ correctly is important. 
Intuitively, it captures the degrees of freedom present in the trajectory data or the number of trajectories that are sufficient to form a basis. 
From the analysis above, we saw that rank $r=4$ corresponds to assuming constant depth and perspective camera.
However, higher $r$ is needed to tolerate changing depth and tracking errors~\cite{Hartley2004,costeira1998multibody}. 
Similarly, not all motion is rigid in real-world videos, which also requires increasing $r$. 
We empirically determined $r=5$ to yield good results.

\begin{figure}
    \centering
    \includegraphics[width=0.95\textwidth, trim={0 3cm 0 0}]{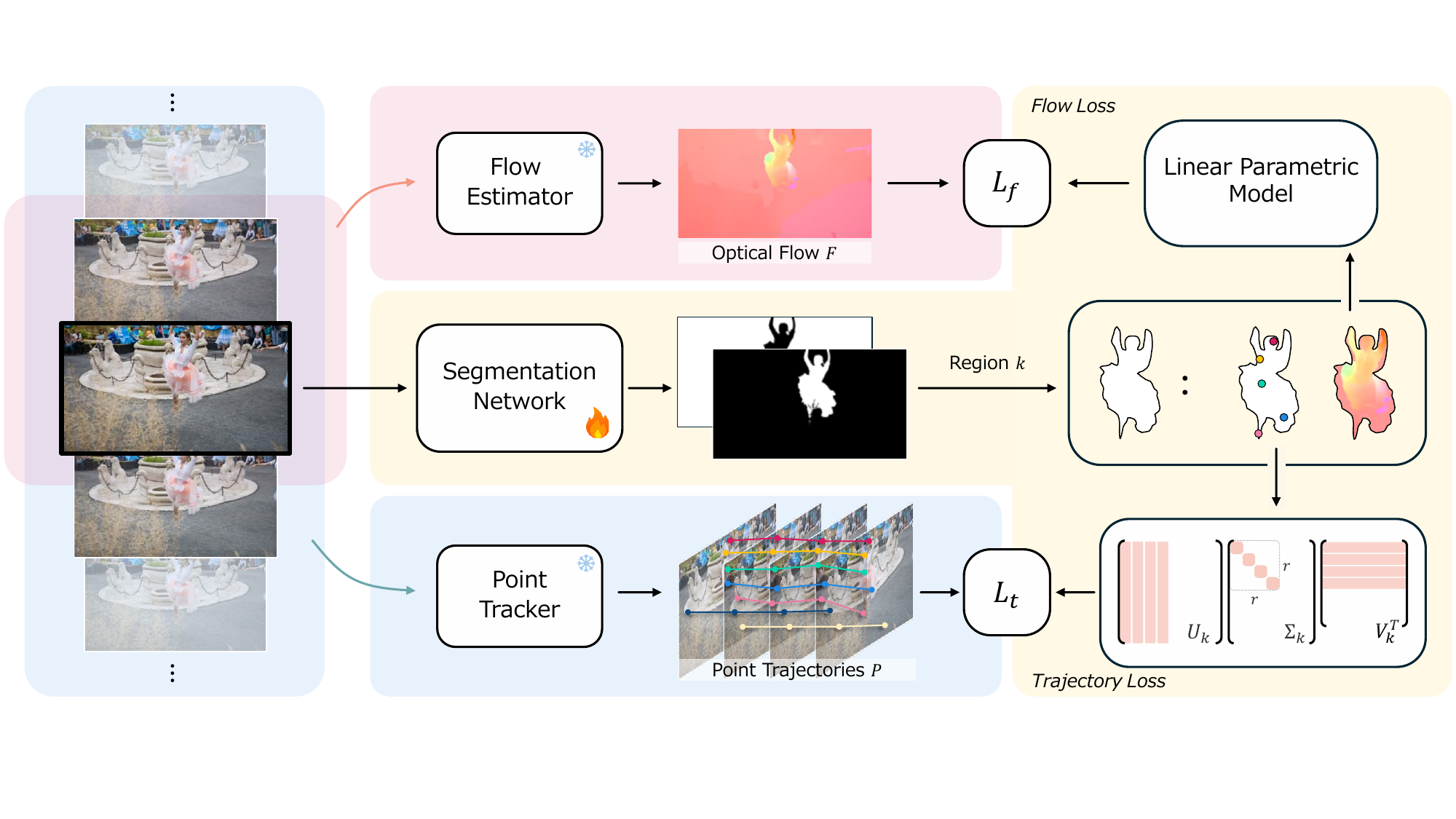}
    \caption{\textbf{Overview of our approach.} We self-supervise a segmentation network, \ie without access to mask annotations, using both short-term motion information (optical flow) and long-term motion (point trajectories). We design a loss function that encourages the segmentation network to cluster regions where trajectories form low-rank-$r$ groups, which should align well with objects. Off-the-shelf methods are used to estimate optical flow and point trajectories given a dataset of videos.
    \label{fig:method}
    }
\end{figure}
\subsection{Training a segmenter using flow and trajectories}

The losses above require optical flow $F$, trajectories $P$, and masks $M_k$ obtained using a segmentation network $\Phi(\mathcal{I}) = M$. This suggests a simple procedure of training a segmentation network given a dataset of videos, which we summarise in \cref{fig:method}. We precalculate optical flow for each frame and obtain a set of point trajectories for each video using off-the-shelf pretrained networks.
For training, we consider triples of $(\mathcal{I},F,P)_i$ for each frame $i$, where for trajectories $P$, we take trajectories for which the points are visible in the image $\mathcal{I}$. This can be accomplished by making use of visibility predictions in the output of point trackers or calculating trajectories by querying points in each frame. We use bilinear sampling for $\pi(\cdot)$ to obtain mask values at trajectory coordinates.

\paragraph{Temporal smoothing.} We include a temporal smoothing loss, which matches mask predictions between two frames offset by $\Delta t$ using the predicted trajectories:
\begin{equation}
    \mathcal{L}_\tau = \| \pi(\Phi(\mathcal{I}_{t}), (P_t)_t)  - \pi(\Phi(\mathcal{I}_{t+\Delta t}), (P_t)_{t +\Delta t})\|^2_2\,,
\end{equation}
where $\mathcal{I}_{t}$ is the $t$-th frame and $P_t$ are trajectories associated with $t$-th frame. 
We write the final loss as:
$    \mathcal{L} = \lambda_f \mathcal{L}_f + \lambda_t \mathcal{L}_t +\lambda_\tau \mathcal{L}_\tau, 
$
where $\lambda_f,\lambda_t,\lambda_\tau$ balance the contribution of the different loss terms. %

\paragraph{Choice of $k$.} Following prior work~\cite{choudhury+karazija22gwm}, we set $k$, the number of predicted masks, to be higher than the maximum number of objects in the scene to account for potential parallax and non-rigid motion.
In the binary segmentation case, we recover two components by considering the average appearance feature of each component and solving for the normalised cut on a graph with $k$ nodes.

\section{Feasibility study}
Our proposed trajectory loss \eqref{eq:traj_loss} enables training a segmentation network using trajectory data.
We first show the feasibility of the proposed cost function in a controlled setting, without actually training $\Phi$.
To this end, we consider a synthetic scene from the MOVI-F Kubric~\cite{greff2022kubric} dataset for which we obtain ground-truth trajectories for every point and ground-truth object segmentation masks. 
We explore the loss landscape of the proposed formulation by corrupting the segmentation masks along several principled axes and studying the effect of such corruptions on the trajectory loss.

\begin{figure}
\begin{minipage}{.07\textwidth}
  \centering
  \includegraphics[width=\linewidth]{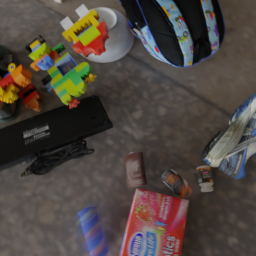}
  \centering
  \includegraphics[width=\linewidth]{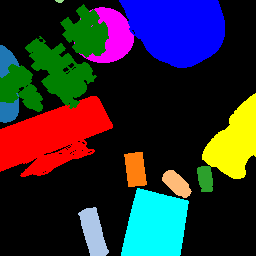}
  \centering
  \includegraphics[width=\linewidth]{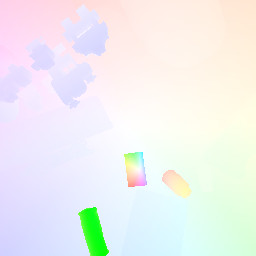}
\end{minipage}
\begin{minipage}{.30\textwidth}
  \centering
  \includegraphics[width=\linewidth]{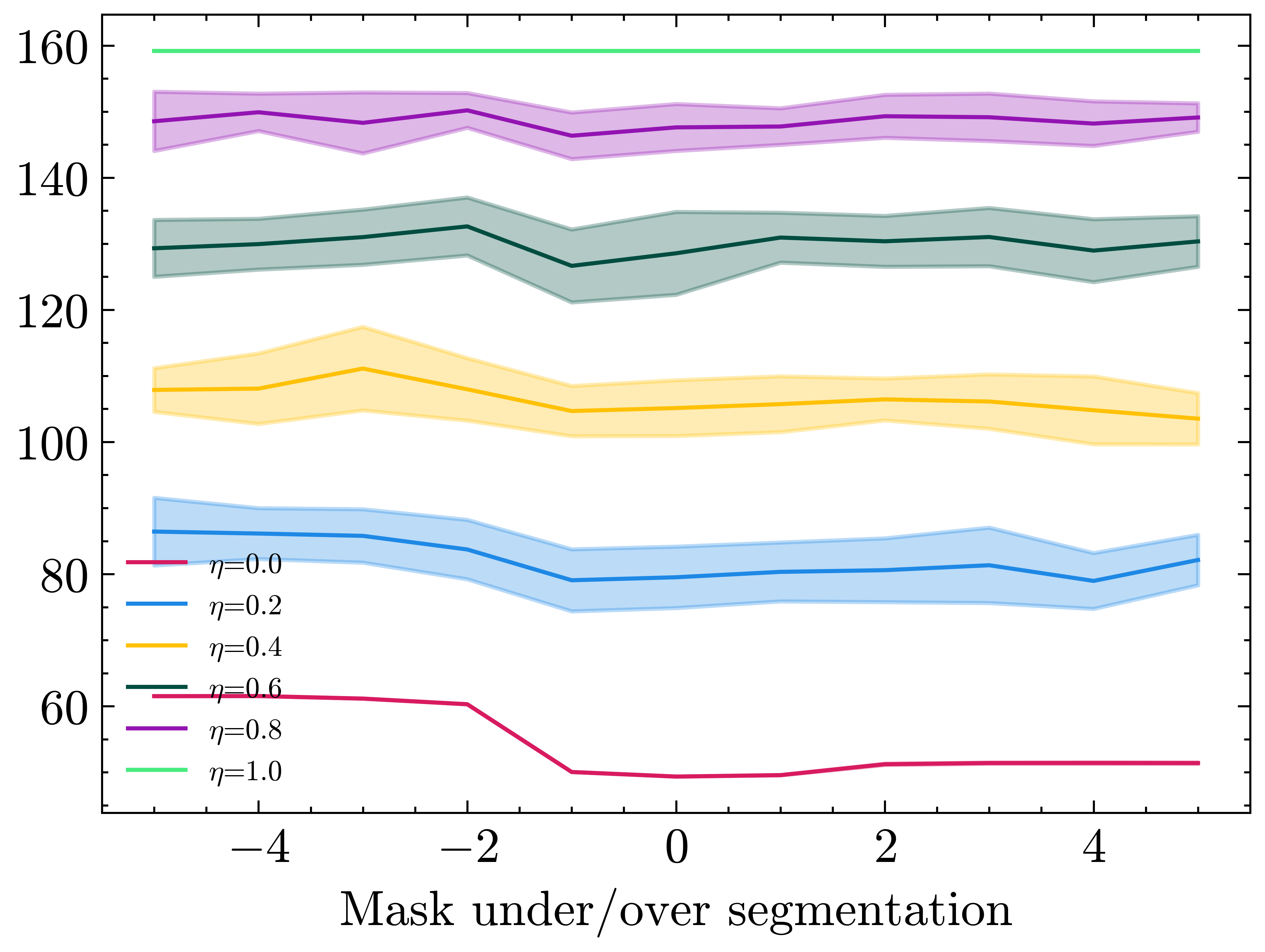}
\end{minipage}%
\begin{minipage}{.30\textwidth}
  \centering
  \includegraphics[width=\linewidth]{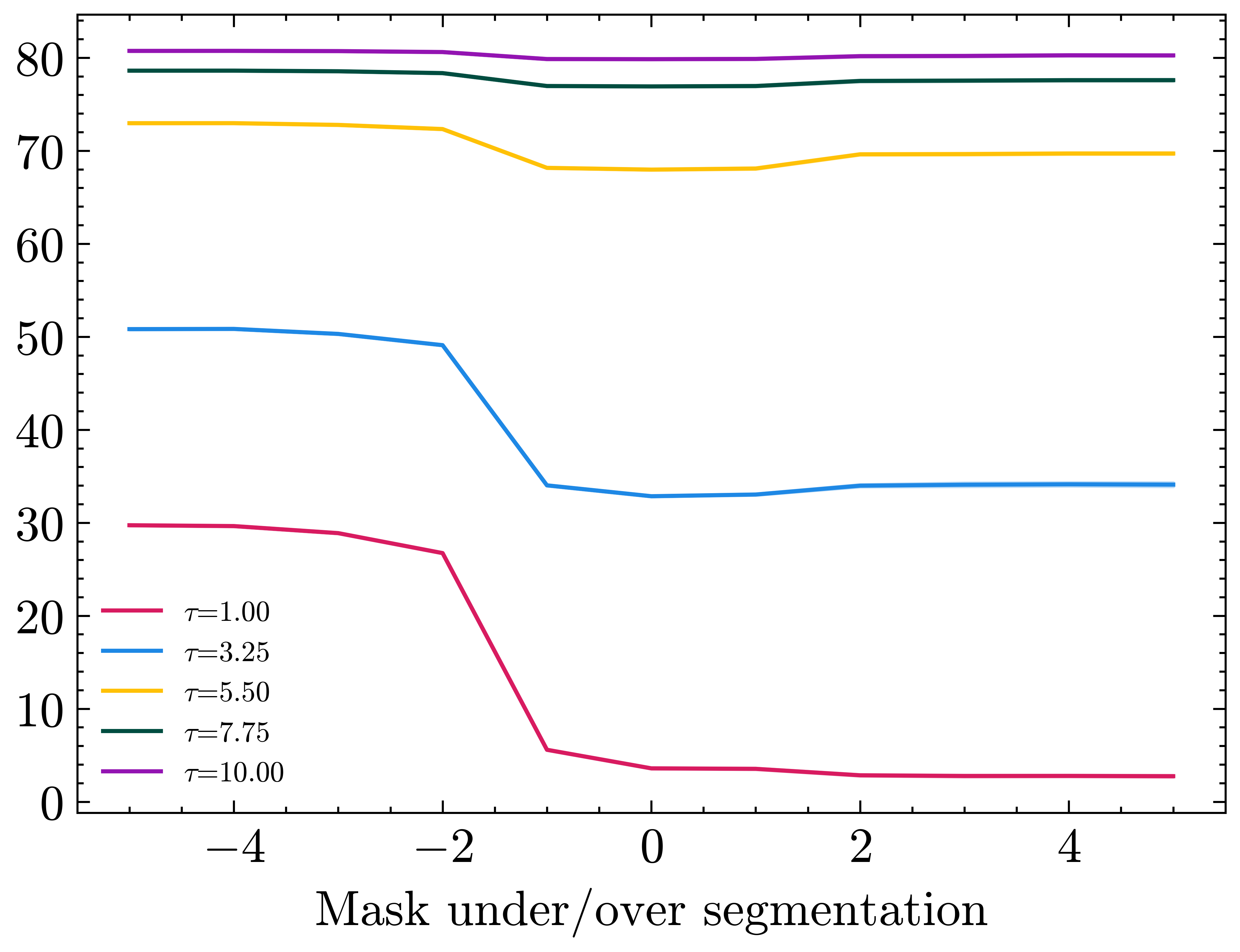}
\end{minipage}%
\begin{minipage}{.30\textwidth}
  \centering
  \includegraphics[width=\linewidth]{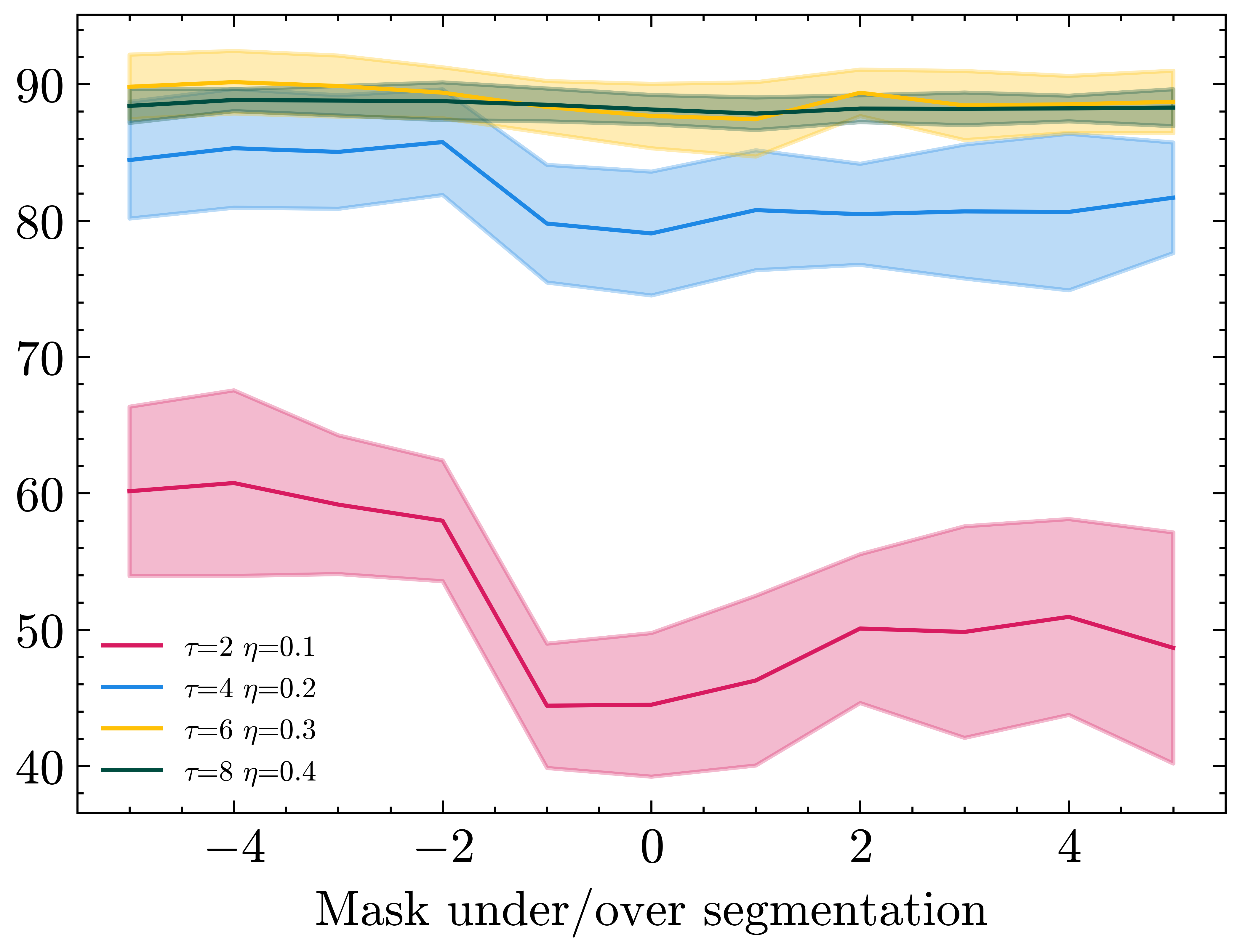}
\end{minipage}
\caption{\textbf{Feasibility analysis of $\mathcal{L}_t$.} 
Using a synthetic sequence (left), we vary the amount of noise $\eta$ injected into the mask, the temperature $\tau$ of the mask logits and plot the loss value as a function of the mask under/over segmentation. The plots show that the loss is reduced in low-noise, low-entropy settings and penalises both over- and under-segmentation. 
}
\label{fig:loss_exploration}
\end{figure}

First, we consider a random alteration of mask pixels, which we refer to as \emph{mask noise}. 
We control the amount of mask noise using $\eta$ such that 0.0 corresponds to no pixels changed and 1.0 corresponds to completely random masks. 
Along this axis, we test whether our loss favours predictions with lower noise. %
Second, we consider structural alterations, namely under/over-segmentation. 
To simulate under-segmentation, we merge object masks with the background at random. 
To simulate over-segmentation, we randomly split the existing object mask into two parts in the middle along either the $x$ or $y$-axis. 
We represent this type of mask corruption using integers.
Negative values indicate the number of objects removed, while positive values correspond to new objects generated from existing ones.
Such structural corruption investigates whether the loss can correctly identify the number of moving objects. 
Finally, we consider the ``softness'' of the predicted masks by transforming masks into logits and increasing the temperature $\tau$ in the softmax operation. 
This tests whether the loss will prefer low-entropy values. 
We leave further details of the corruption procedure to \cref{sec:app_masks}.
The results of these analyses are shown in \Cref{fig:loss_exploration}. 
All three plots show the loss value as a function of structural corruption. 
The trajectory loss decreases as the noise and temperature of the masks are reduced, as seen in the first two plots. 
The third plot also shows that such solutions are preferred in combination. 

Furthermore, we observe that the loss values are lower when the correct number of segments is detected, and this holds even in the presence of noise or when masks are more uniform. 
Note, however, that over-segmentation is penalised less than under-segmentation, \ie missing moving objects leads to a higher value of the loss than, \eg splitting an object into several components.

\section{Experiments}\label{sec:experiments} 
In this section, we evaluate our approach for unsupervised motion segmentation and compare it with simple baselines and prior subspace clustering methods.
Next, we compare our method with state-of-the-art methods for unsupervised video object segmentation across several datasets in a binary segmentation setting.
We finish with ablation experiments of our approach.

\paragraph{Datasets.} We consider four primary datasets in this study. 
We use the synthetic MOVi-F variant of the Kubric~\cite{greff2022kubric} dataset with ground truth trajectories for comparison with subspace clustering-based approaches.
We adopt this setting to eliminate noise in point trajectories as previous methods are sensitive to it.  %
We report the adjusted Rand index (ARI) as the main metric, measuring how close clustering is to the ground truth up to the permutation of cluster identities, where 1 is a perfect match, and 0 means roughly random assignment. We also report FG-ARI, \ie ARI only on foreground pixels (determined by ground truth masks), which identifies how well different objects are separated.

We also evaluate our approach on real-world datasets: DAVIS 2016~\cite{perazzi2016a-benchmark-davis}, SegTrackv2 (STv2)~\cite{li2013video-segtrack}, and FBMS~\cite{ochs2014segmentation-fbms}, which are popular benchmarks for video object segmentation. 
Following standard practice~\cite{yang2021self-supervised,yang-loquercio2019unsupervised}, foreground objects in STv2 and FBMS are consolidated. 
We report the Jaccard ($\mathcal{J}$) score, computed using Hungarian matching between predicted and ground truth segmentations.

\paragraph{Implementation.} 

\begin{wraptable}[12]{r}{0.35\linewidth}
\vspace{-17pt}

\begin{center}
\footnotesize

\caption{
Comparison of our LRTL trajectory-based formulation with prior methods. 
}
\begin{tabular}{@{}lcc@{}}
\toprule
 & \multicolumn{2}{c}{\textbf{MOVi-F}} \\
\textbf{Method} & ARI$\uparrow$ & FG-ARI$\uparrow$  \\
\midrule
K-Means & 15.26 & 42.53 \\
SSC~\cite{elhamifar2013sparse} & 11.12 & 39.21 \\
LRR~\cite{liu2012robust} & 7.47 & 37.36 \\
\midrule
\textbf{LRTL (Ours)} & 46.07 & 65.76 \\
\bottomrule
\end{tabular}
\label{tab:track_opt}

\end{center}

\end{wraptable}

For the experiments on real-world datasets, optical flow is estimated using RAFT~\cite{teed2020raft} and point trajectories using CoTracker~\cite{karaev2023cotracker}.
Trajectories are computed within a context window $f=20$ around each frame, with reflection padding around video boundaries, resulting in chunks of $T = 2f+1=41$ frames.
To reduce the effect of noisy predictions, we also filter trajectories along the time dimension using an average filter with a window size of 11.
For the experiments on MOVi-F, a small U-Net \cite{ronneberger2015u} is trained as the segmentation network, starting from random initialisation.
For fairness of comparisons on DAVIS, STv2 and FBMS, we use the same architecture as in~\cite{choudhury+karazija22gwm}\,---\,MaskFormer with DINO backbone. We specify further details in \cref{sec:details}.

\subsection{Comparison to trajectory-based methods}

In \cref{tab:track_opt}, we compare our \underline{l}ow-\underline{r}ank \underline{t}rajectory \underline{l}oss (LRTL) with prior subspace clustering approaches in a per-video optimisation setting.
Subspace clustering operates on a similar intuition to our proposed trajectory loss by a grouping of trajectories that should be linearly dependent.
We also consider K-means clustering of trajectories as a simple baseline.
For fair comparisons, we train our segmentation model optimising \emph{only} the trajectory loss ($\mathcal{L}_t$).
We use $k=25$ components for each video and train for 5000 steps. 
This is comparable to the computation requirements and steps of other methods.
For  K-means, SSC~\cite{elhamifar2013sparse} and LRR~\cite{liu2012robust}, we search for an optimal set of hyperparameters and the number of components $k$, reporting the best results. 
Our approach shows significantly stronger performance than simple K-Means and subspace clustering approaches.

\begin{table}[t]

\footnotesize
\begin{center}
\caption{
\textbf{Unsupervised video segmentation} on DAVIS, SegTrackv2, and FBMS. 
Where possible, we report results without widely applicable post-processing (\eg CRF) or indicate results in \textcolor{gray}{grey}.
}
\vspace{5pt}
\begin{tabular}{lccccccc}
\toprule
& \multicolumn{2}{c}{Inf. Input} & Input & Motion Est. & \textbf{DAVIS} & \textbf{STv2} & \textbf{FBMS} \\
\textbf{Method} & RGB                       & Motion    & Resolution            & Method &  $\mathcal{J}\uparrow$ & $\mathcal{J}\uparrow$ & $\mathcal{J}\uparrow$ \\
\midrule

\multicolumn{3}{l}{\textit{Single-sequence methods}} \\ 
FTS~\cite{papazoglou2013fast}                  & \cmark & \cmark & --               & LDOF~\cite{brox2010ldof} & 55.8 & 47.8 & 47.7\\
CUT~\cite{keuper2015motion}                   & \cmark & \cmark & --               & LDOF~\cite{brox2010ldof} & 55.2 & 54.3 & 57.2\\
DS~\cite{ye2022deformable}   & \cmark & \cmark & $240 \times 426$ & RAFT~\cite{teed2020raft} & 79.1 & 72.1 & 71.8\\
\color{gray}Ponimatkin et al.~\cite{ponimatkin2023simple} & \color{gray}\cmark & \color{gray}\xmark & \color{gray}$480 \times 848$ & \color{gray}ARFlow~\cite{liu2020learning} & \color{gray}80.2 & \color{gray}74.9 &\color{gray} 70.0\\
\color{gray}OCLR~\cite{Xie2022SegmentingMO} (test ft.)         & \color{gray}\cmark & \color{gray}\cmark & \color{gray}$480 \times 848$ & \color{gray}RAFT~\cite{teed2020raft} & \color{gray}80.9 &\color{gray}72.3 & \color{gray} 69.8\\
\midrule

\multicolumn{3}{l}{\textit{Single-stage end-to-end methods}} \\
OCLR~\cite{Xie2022SegmentingMO}         & \xmark & \cmark & $112 \times 224$ & RAFT~\cite{teed2020raft} & 72.1 & 67.6 & 65.4\\
DivA~\cite{lao2023divided}                  & \cmark & \cmark & $128 \times 224$ & RAFT~\cite{teed2020raft} & 72.4 & 64.6 & 60.9\\
Meunier et al.~\cite{Meunier_2023_CVPR}      & \xmark & \cmark & $128 \times 224$ & RAFT~\cite{teed2020raft} & 73.2 & 55.0 & --\\
GWM~\cite{choudhury+karazija22gwm}                 & \cmark & \xmark & $128 \times 224$ & RAFT~\cite{teed2020raft} & 79.5 & 78.9 & 78.4\\
\midrule

\multicolumn{3}{l}{\textit{Multi-stage methods}} \\
RCF~\cite{lian2023bootstrapping}   & \cmark & \xmark & $480 \times 848$ & RAFT~\cite{teed2020raft} & 80.9 & 76.7 & 69.9\\
LOCATE~\cite{singh2023locate} & \cmark & \xmark & $480 \times 848$ & ARFlow~\cite{liu2020learning} & 80.9 & 79.9 & 68.8\\

\midrule
\multirow{2}{*}{\textbf{LRTL (Ours)}}            & \multirow{2}{*}{\cmark} & \multirow{2}{*}{\xmark} & \multirow{2}{*}{$192 \times 352$} & RAFT~\cite{teed2020raft} & \multirow{2}{*}{\textbf{82.2}} & \multirow{2}{*}{\textbf{81.2}} & \multirow{2}{*}{\textbf{79.6}} \\
                                         &                         &                         &                                   & CoTracker~\cite{karaev2023cotracker}\\

\bottomrule
\end{tabular}
\label{tab:main}
\end{center}
\end{table}
\subsection{Unsupervised video object segmentation}
We compare to recent methods on the unsupervised video object segmentation task \textit{without first-frame prompting or post-processing}. 
In this setting, we train a single network on the benchmark datasets for binary video segmentation.
We compare with \textit{single-sequence methods} that perform optimisation for each sequence/video individually. 
Additionally, we benchmark dataset-wide \textit{single-stage end-to-end methods} where training is performed over multiple videos simultaneously, training a network in an end-to-end manner.
We also compare with \textit{multi-stage methods} that train and re-train several networks.
We report our results on standard benchmarks in \cref{tab:main}.
While the closest prior work relies on multiple stages of training, pseudo-labelling, applying CRF, and retraining, our end-to-end trained method shows better performance at lower resolutions.
We attribute this to the effectiveness of our approach in incorporating long-term motion information.

\begin{figure}[th]
    \centering
    \includegraphics[width=\textwidth,trim={0.5cm 0cm 0cm 0cm}]{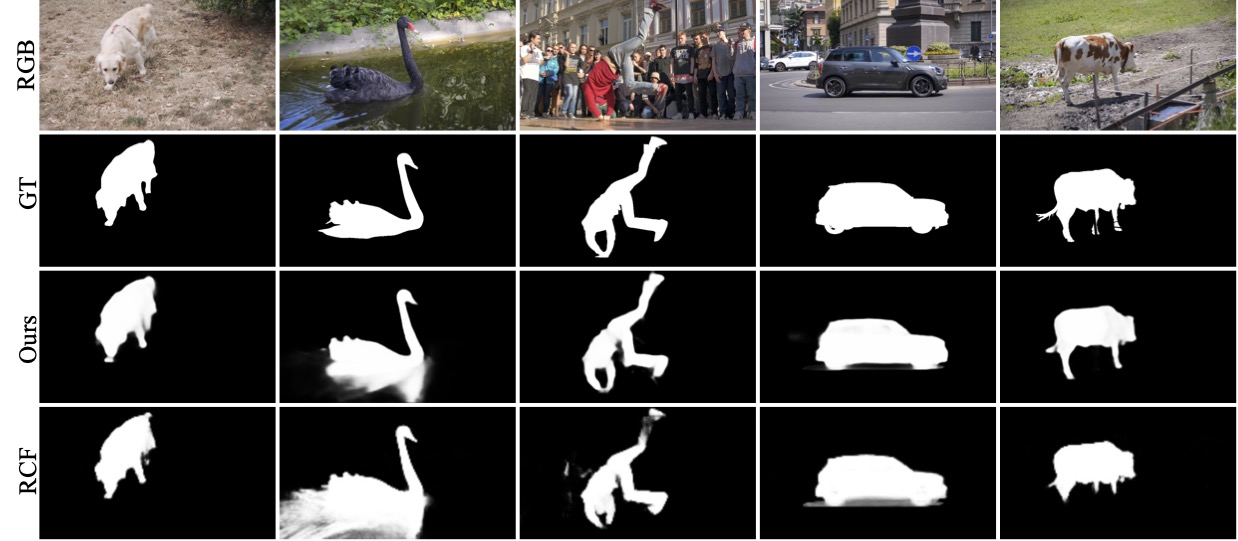}
    \caption{
    Qualitative comparison of our results on DAVIS with RCF which uses higher resolution and multi-stage training. 
    Our method contains slightly better boundaries, does not segment shadows and separates water ripples from the swan.
    }
    \label{fig:qual}
\end{figure}

In \cref{fig:qual}, we show qualitative results of our approach and compare with RCF~\cite{lian2023bootstrapping}, a state-of-the-art multi-stage approach. 
Our network trained with both flow and trajectory losses yields segmentations with noticeably better boundaries despite operating at a lower resolution. 
Notably, our formulation also effectively avoids segmenting shadows and water ripples of the swan, which are difficult to separate based on instantaneous motion alone. 

\begin{table}
\begin{minipage}[t]{0.48\textwidth}%
\centering
\footnotesize
\caption{
\textbf{Alternative losses} to our proposal.
Other variants do not match the performance of our formulation.
}%
\vspace{1pt}
\label{tab:alt_losses}
\begin{tabular}{cc}
\toprule
\textbf{Loss} &  \textbf{DAVIS} ($\mathcal{J}\uparrow$) \\
\midrule
$\mathcal{L}_{\mathrm{rec@}3}$ \eqref{eq:loss_rec} & 11.1 \\
$\mathcal{L}_\mathrm{per}$ \eqref{eq:loss_per} & 18.2 \\
$\mathcal{L}_{\mathrm{rec@}5}$ \eqref{eq:loss_rec} & 14.6 \\
\textit{tracks-as-flow} & 65.3 \\ 
\midrule
\textbf{Ours} $\mathcal{L}_t$ \eqref{eq:traj_loss} & 71.9 \\
\bottomrule
\end{tabular}
\vspace{-10pt}%
\end{minipage}%
\hfill
\begin{minipage}[t]{0.48\textwidth}%
\centering
\footnotesize
\caption{
\textbf{Ablation of loss terms.}
All loss terms synergise to improve performance.
} %
\vspace{10pt}
\label{tab:loss_abl}
\begin{tabular}{cc}
\toprule
 \textbf{Loss} & \textbf{DAVIS} ($\mathcal{J}\uparrow$) \\
\midrule
$\lambda_f \mathcal{L}_f$ &  78.5 \\
$\lambda_t \mathcal{L}_t$ &  71.9 \\
$\lambda_t \mathcal{L}_f + \lambda_t \mathcal{L}_t$ & 81.7 \\
\midrule
$\lambda_t \mathcal{L}_f + \lambda_t \mathcal{L}_t + \lambda_\tau \mathcal{L}_{\tau}$ & 82.2 \\
\bottomrule
\end{tabular}
\vspace{-10pt}%
\end{minipage}%
\end{table}
\subsection{Ablations}

\paragraph{Alternative losses.}\label{sec:alt_losses}
We have explored several alternative formulations of the trajectory loss in our approach and present the analysis in \cref{tab:alt_losses}.
Losses based on full SVD reconstruction fail to train a network sufficiently.
$\mathcal{L}_\mathrm{per}$ performs the best out of these, likely as DAVIS contains several scenes with a panning camera tracking a rigid object at an approximately constant distance, which matches the assumptions.
Increasing or decreasing the rank of the approximation performs worse.
We also consider \textit{track-as-flow} loss, where trajectories $P$ are treated as optical flow by subtracting positions from adjacent times.
Then, for $T$ frames, \cref{eq:flow_loss} can be applied.
We find that such a formulation still underperforms in comparison to our trajectory-based formulation (\cref{eq:traj_loss}).

We believe our formulation provides better results than the above for two possible reasons. 
First, by minimising higher-than-$r$ singular values, we are not \emph{strictly} enforcing assumptions like rigidity. 
Second, our loss formulation is more numerically stable as it requires only gradients w.r.t. to the singular values. As we seek to drive them close to zero, the matrices $P_k$ become increasingly ill-conditioned as the training progresses. Additionally, gradients w.r.t. $U$ and $V^\top$ depend on inverse singular values $\Sigma^{-1}$~\cite{townsend2016svd}, which become numerically unstable as they are approaching zero. On the other hand, $\mathop{\operatorname{d}\Sigma} = I_N \circ (U^\top\mathop{\operatorname{d}P_k}V)$ does not have this problem. 

\paragraph{Influence of losses.} In \cref{tab:loss_abl}, we consider the method with only the flow loss component and only the trajectory loss component. We find that our trajectory-based loss improves flow-only performance. Using only trajectory-based loss shows weaker performance than just optical flow, likely due to using only a sparse set of points and the noise introduced by estimating positions for occluded points. Ablating temporal smoothing loss slightly lowers performance as well.

\paragraph{Limitations.}\label{sec:limitations}
While we have demonstrated the effectiveness of learning segmentation from long-term motion, there is potential for further improvements in leveraging point trajectories.
First, while modern trackers predict reasonable positions for occluded points, naturally, these predictions are less accurate. Thus, a more explicit handling of occlusions and tracking noise would likely help.
Second, we currently only use trajectory estimates from nearby frames for training.
This means that we sometimes track the same point multiple times, which could be avoided with caching trajectories.
While we handle non-rigidity using over-segmentation, extending this principle to video with multiple non-rigid objects is an important feature direction.
\section{Conclusion}\label{sec:conclusion}

We have introduced a principled method to train an image segmentation network using long-term motion information expressed as point trajectories. Our trajectory loss formulation follows the principle of common fate and aims to group trajectories into low-rank matrices, representing the idea the motion of points belonging to the same object can be roughly explained as a combination of other points. Using synthetic data we have shown that such a loss should prefer low-noise and low-entropy solutions as well as identify the correct number of moving objects. In comparison with other methods, our loss formulation has shown superior performance compared to subspace clustering baselines on synthetic data and achieved state-of-the-art results on unsupervised video object segmentation benchmarks when combined with optical flow-based loss.

\paragraph{Acknowledgements}
L.~K. is supported by is supported by EPSRC AIMS CDT EP/S024050/1.
I.~L., C.~R. and A.~V. are supported by ERC-CoG UNION 101001212 and EPSRC VisualAI EP/T028572/1.

{
\small
\bibliographystyle{plainnat}
\bibliography{main}
}

\newpage
\appendix
\section*{Supplementary material}

In this supplementary material, we consider additional ablations of our approach (\cref{sec:app_abla}), include further results (\cref{sec:app_res}), and provide the implementation details (\cref{sec:details}). Accompanying this supplementary material, we include videos of our results on DAVIS and SegTrackv2 datasets. We also include an example video visualising a sample of trajectories that the model receives as input. The code and models will be released upon acceptance.

\section{Broader impact}\label{sec:impact}
Segmentation is a component in a very large and diverse spectrum of applications in healthcare, image processing, computer graphics, surveillance and more. As with many technologies, the application can be good or bad. In this paper, we explore how to train a model to perform segmentation in an unsupervised manner. This has the positive benefit of removing manual labour requirements to obtain annotations, which might also eventually apply to bad actors. We, however, consider the immediate real-world impact beyond the research community of our work here limited as unsupervised systems still show lower performance than supervised counterparts.

\section{Additional ablations}\label{sec:app_abla}

\begin{table}

\begin{minipage}{0.48\textwidth}

\begin{center}
\footnotesize

\caption{
\textbf{Influence of $r$}, the rank of the trajectory matrix used in loss function \eqref{eq:traj_loss}.
}
\begin{tabular}{cc}
\toprule
$r$ & \textbf{DAVIS} ($\mathcal{J}\uparrow$) \\
\midrule
3 & 76.0 \\
4 & 79.6 \\
5 & 82.2 \\
6 & 80.9 \\
\bottomrule
\end{tabular}
\label{tab:dof_abl}

\end{center}

\end{minipage}
\hfill
\begin{minipage}{0.48\textwidth}

\begin{center}
\footnotesize

\centering
\caption{
\textbf{Influence of $k$}, the number of predicted components before merging.
}
\begin{tabular}{cc}
\toprule
k & \textbf{DAVIS} $\mathcal{J}\uparrow$ \\
\midrule
2 & 78.0 \\
3 & 82.0 \\
4 & 82.2 \\
5 & 72.8 \\
\bottomrule
\end{tabular}
\label{tab:k_abl}
    
\end{center}

\end{minipage}
\hfill
\begin{minipage}{0.48\textwidth}

\begin{center}
\footnotesize
    
\caption{
\textbf{Influence of context length} of the trajectory matrix. 
}
\begin{tabular}{cc}
\toprule
Context length &  \textbf{DAVIS} ($\mathcal{J}\uparrow$) \\
\midrule
10 & 79.1 \\
15 & 81.0 \\
20 & 82.2 \\
30 & 80.8 \\
\bottomrule
\end{tabular}
\label{tab:context_abl}

\end{center}

\end{minipage}
\hfill
\begin{minipage}{0.48\textwidth}

\begin{center}
\footnotesize

\caption
{
\textbf{Influence of trackers} used to estimate point trajectories.
}
\begin{tabular}{cc}
\toprule
\textbf{Tracker} &  \textbf{DAVIS} ($\mathcal{J}\uparrow$) \\
\midrule
TAPIR~\cite{doersch2023tapir} & 73.4  \\
PIPs++~\cite{zheng2023point} &  74.9 \\
BootsTap~\cite{doersch2024bootstap} & 76.8 \\ 
CoTracker~\cite{karaev2023cotracker} & 78.9 \\
\bottomrule
\end{tabular}
\label{tab:tracker_abl}

\end{center}

\end{minipage}

\end{table}

\paragraph{Rank $r$ of trajectory matrix.} In \cref{tab:dof_abl}, we vary $r$, the rank of the trajectory matrix used in the trajectory loss (\cref{eq:traj_loss}). As previously mentioned, the choice of rank reflects the degrees of freedom in the system and controls implicitly the assumptions about the types of motion and cameras used to capture sequences. At $r=3$ and 4, we observe slight impact on the performance in comparison to $r=5$. $r=5$ appears to be the optimal setting, which is what we used in our main experiments. At $r=6$, the performance drops again, likely as it becomes sufficient to group and explain simple motions together. 

\paragraph{Number of segments $k$.} In \cref{tab:k_abl}, we vary k, the number of masks predicted by our method, before merging. As in prior work~\cite{choudhury+karazija22gwm}, the $k=4$ appears to be the optimal setting. The performance drops beyond this point as it becomes difficult to group objects.

\paragraph{Influence of context window length $T$.} In \cref{tab:context_abl}, we vary the length of the context windows ($f$) and thus $T$ for our method when considering trajectories. We find increasing the context window helps slightly. However, the performance starts to drop afterwards. We hypothesise that this is due to difficulty predicting sensible point trajectories for points that move outside of the frame and become invisible, as DAVIS contains many videos where the camera tracks the main subject. Though several values of this setting are viable.

\paragraph{Source of tracks.} 
In \cref{tab:tracker_abl}, we experiment with different trackers to obtain tracks. 
We consider TAPIR\footnote{Code and models available \url{https://github.com/google-deepmind/tapnet} under Apache-2.0 license.}, PIP++\footnote{Code and models available \url{https://github.com/aharley/pips2} under MIT license.} and BootsTap\footnote{Code and models available \url{https://github.com/google-deepmind/tapnet} under Apache-2.0 license.} along CoTracker.
Due to the limitation of some options (PIPs++ not predicting visibility) and inherent noise in invisible tracks for TAPIR, we lowered the context window to 15. We also do not consider tracks from adjacent frames as this seems to lower performance for other trackers. Finally we did not use EMA in these experiments.
We observe that CoTracker performs the best while other trackers show slightly weaker results. 
We hypothesise this is due to CoTracker estimating reasonable trajectories for occluded points, which are included in the matrix $P_k$. 
Some trackers, \eg TAPIR, are restricted to predicting points within the frame, thus providing extremely noisy estimates in scenes where objects move outside the frame.

\paragraph{Alternative networks.}
As our proposed loss function is network-architecture agnostic as it only requires mask prediction. 
Thus, any network which predicts masks or has mask-like representation could be used. 
In \cref{tab:networks}, we experiment with changing the segmenter architecture in the DAVIS benchmark. This shows that we can swap different network architectures with relative ease and obtain similar results.

\paragraph{Inference speed.}
Here we provide the inference time comparison using different networks as average FPS during DAVIS evaluation. For MaskFormer + DINO configuration, we measure 3.3 FPS, while with UNet we measure 6.4 FPS. Note that since our contribution is a loss function, it is network architecture agnostic. Using it does not affect inference time; only the choice of network architecture does. We matched the architecture with prior work for the best comparisons.

\paragraph{Comparison with appearance-only works.}
Finally, we include a comparison to unsupervised methods that consider only appearance during learning. In 
\cref{tab:videocutler}, we provide a comparison of VideoCutLER~\cite{wang2023videocutler} and VideoSAUR~\cite{zadaianchuk2024object} on DAVIS using the same merging strategy for combining multiple predictions to a binary segmentation as in our method.

Our method shows a significant advantage. We observe that VideoCutLER has trouble segmenting instances from crowds in the background. VideoSAUR has imprecise object boundaries which severely impacts performance when measurred using Jaccard score.

\begin{table}

\begin{minipage}{0.48\textwidth}

\begin{center}
\footnotesize
    
\caption{
\textbf{Alternative network architectures} for segmentation.
}
\begin{tabular}{cc}
\toprule
Network &  \textbf{DAVIS} ($\mathcal{J}\uparrow$) \\
\midrule
UNet & 80.6 \\
MaskFormer + Swin-Tiny & 81.2 \\
MaskFormer + DINO & 82.2 \\
\bottomrule
\end{tabular}
\label{tab:networks}

\end{center}

\end{minipage}
\hfill
\begin{minipage}{0.48\textwidth}

\begin{center}
\footnotesize
    
\caption{
\textbf{Comparison with appearance-only methods.} 
}
\begin{tabular}{cc}
\toprule
Method &  \textbf{DAVIS} ($\mathcal{J}\uparrow$) \\
\midrule
VideoCutLER~\cite{wang2023videocutler} & 67.2 \\
VideoSAUR~\cite{zadaianchuk2024object} & 17.5 \\
\textbf{LRTL (Ours)} & 82.2 \\
\bottomrule
\end{tabular}
\label{tab:videocutler}

\end{center}

\end{minipage}

\end{table}

\section{Additional results}\label{sec:app_res}

\subsection{Qualitative results on SegTrackv2}
\begin{figure}[t]
    \centering
    \includegraphics[width=\textwidth,trim={0.5cm 0cm 0cm 0cm}]{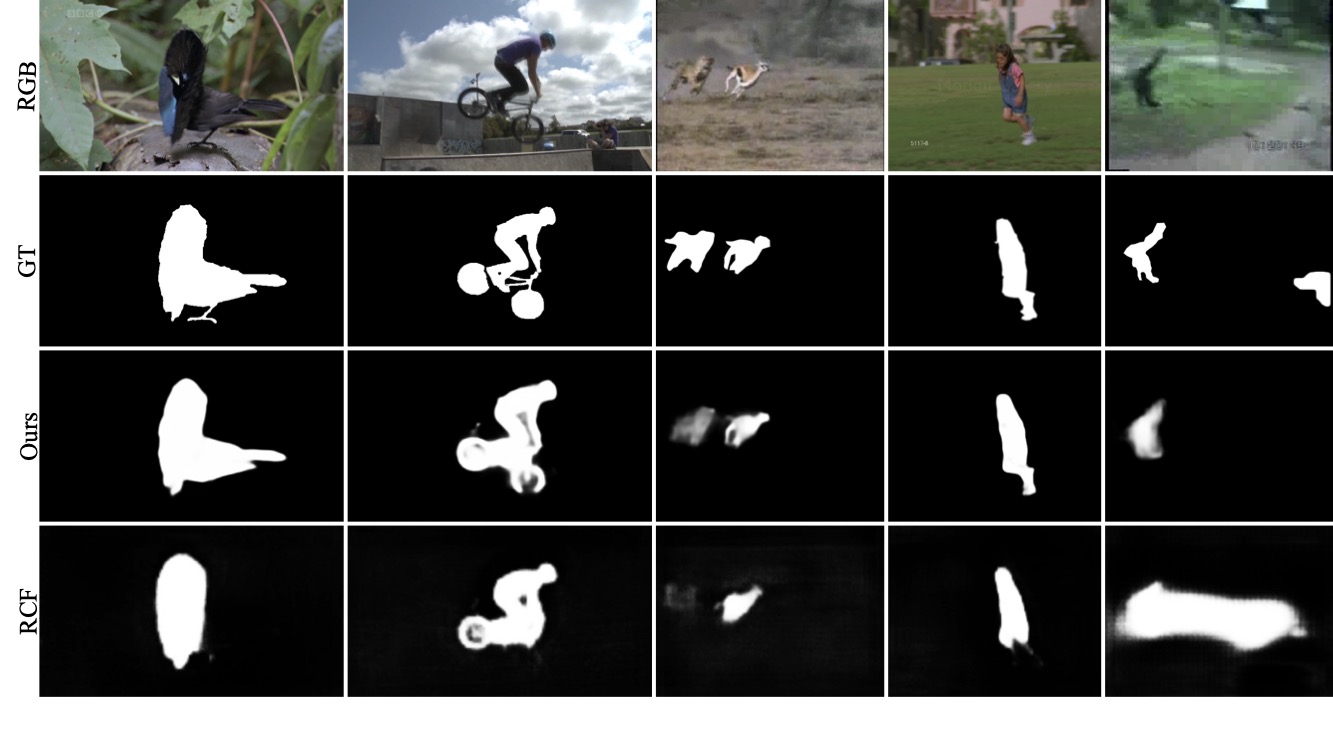}
    \vspace{-2.5em}
    \caption{Qualitative comparison of our results on SegTrackv2 with RCF which uses higher resolution and multi-stage training. Our method contains slightly better boundaries and segments more whole objects.}
    \label{fig:qual_stv2}
\end{figure}
In \cref{fig:qual_stv2}, we provide additional qualitative results from our approach on the SegTrackv2 dataset. We compare with the state-of-the-art multi-stage Relaxed Common Fate (RCF) approach~\cite{lian2023bootstrapping}. Our method correctly identifies more parts of the objects and has better boundaries.

\section{Parametric mask alterations}\label{sec:app_masks}
\begin{figure}[t]
    \centering
    \begin{minipage}[]{0.19\textwidth}
        \includegraphics[width=\textwidth]{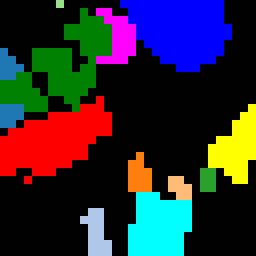}
        \caption*{$\eta=0.0$}
    \end{minipage}
    \begin{minipage}[]{0.19\textwidth}
        \includegraphics[width=\textwidth]{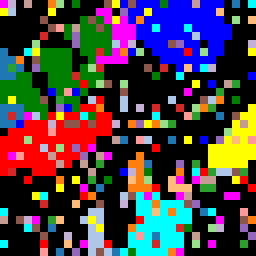}
        \caption*{$\eta=0.25$}
    \end{minipage}
    \begin{minipage}[]{0.19\textwidth}
        \includegraphics[width=\textwidth]{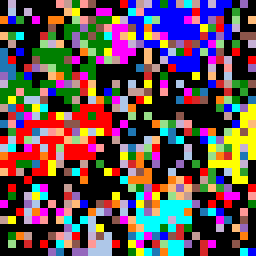}
        \caption*{$\eta=0.5$}
    \end{minipage}
    \begin{minipage}[]{0.19\textwidth}
        \includegraphics[width=\textwidth]{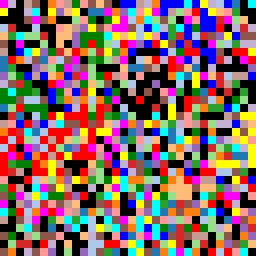} 
        \caption*{$\eta=0.75$}
    \end{minipage}
    \begin{minipage}[]{0.19\textwidth}
        \includegraphics[width=\textwidth]{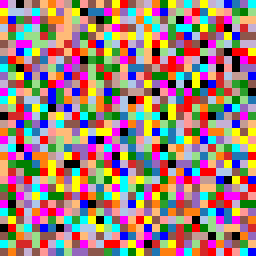} 
        \caption*{$\eta=1.0$}
    \end{minipage}
    \caption{Example \emph{noise} mask alteration. The parameter $\eta$ is the probability of assigning a mask pixel at random.}
    \label{fig:mask_noise_exp}
\end{figure}

\begin{figure}[t]
    \centering
    \begin{minipage}[]{0.19\textwidth}
        \includegraphics[width=\textwidth]{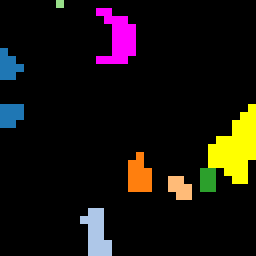}
        \caption*{$s=-4$}
    \end{minipage}
    \begin{minipage}[]{0.19\textwidth}
        \includegraphics[width=\textwidth]{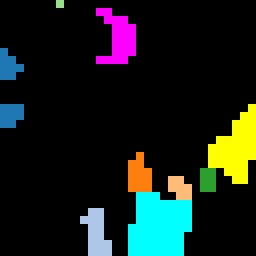}
        \caption*{$s=-3$}
    \end{minipage}
    \begin{minipage}[]{0.19\textwidth}
        \includegraphics[width=\textwidth]{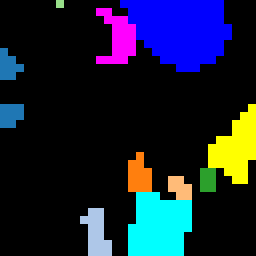}
        \caption{$s=-2$}
    \end{minipage}
    \begin{minipage}[]{0.19\textwidth}
        \includegraphics[width=\textwidth]{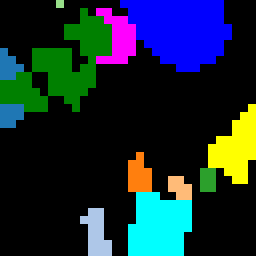} 
        \caption*{$s=-15$}
    \end{minipage}
    \begin{minipage}[]{0.19\textwidth}
        \includegraphics[width=\textwidth]{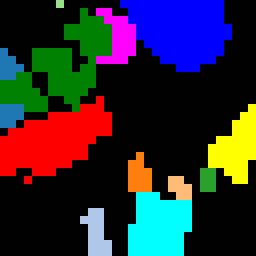} 
        \caption*{$s=0$}
    \end{minipage}
    \caption{Example mask \emph{structural} mask alteration showing modification used to approximate under-segmentation. The parameter $s$ controls the number of objects set to the background.}
    \label{fig:mask_under_exp}
\end{figure}

\begin{figure}[!t]
    \centering
    \begin{minipage}[]{0.19\textwidth}
        \includegraphics[width=\textwidth]{figures/images/mask_struct_5.png}
        \caption*{$s=0$}
    \end{minipage}
    \begin{minipage}[]{0.19\textwidth}
        \includegraphics[width=\textwidth]{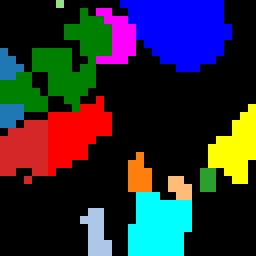}
        \caption*{$s=1$}
    \end{minipage}
    \begin{minipage}[]{0.19\textwidth}
        \includegraphics[width=\textwidth]{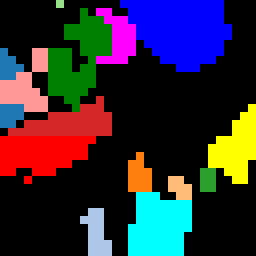}
        \caption*{$s=2$}
    \end{minipage}
    \begin{minipage}[]{0.19\textwidth}
        \includegraphics[width=\textwidth]{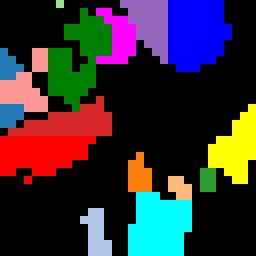} 
        \caption*{$s=3$}
    \end{minipage}
    \begin{minipage}[]{0.19\textwidth}
        \includegraphics[width=\textwidth]{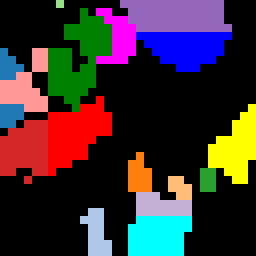} 
        \caption*{$s=4$}
    \end{minipage}
    \caption{Example \emph{structural} mask alteration modelling over-segmentation. The parameter $s$ controls the number if objects split into two at random.}
    \label{fig:mask_over_exp}
\end{figure}

In this section, we show the effect of the parametric ground truth mask alterations used to study the trajectory loss in section 4.1. 
The purpose of these alterations is to disturb ground truth masks in a controlled way to enable studying the effect this has on the loss. 
For this purpose, we use synthetic data from MOVi-F sequences of the Kubric~\cite{greff2022kubric} dataset suite, which is the same data that is used to train CoTracker~\cite{karaev2023cotracker}, TAPIR~\cite{doersch2023tapir}, PIP~\cite{harley2022particle} and similar.
We consider three types of alterations:
\begin{itemize}
    \item The first kind of alteration is random \emph{noise}. With probability $\eta$, we set each mask pixel to a random class sampled from $\mathcal{U}(0, K)$, where $K=20$ in this case. When $\eta=0$, thus, there is no alteration. When $\eta=0.5$, around half of the mask pixels (in expectation) are assigned randomly. \cref{fig:mask_noise_exp} shows the effect of $\eta$ in practice.
    \item The second kind of alteration we consider is a \emph{structural} change meant to approximate over/under-segmentation. For under-segmentation, we change the mask regions corresponding to the whole object to the background. \cref{fig:mask_under_exp} shows this in effect. For over-segmentation, we split an existing component randomly along an axis passing through the object centre and parallel to either the x- or y-axis at random. \cref{fig:mask_over_exp} shows this in effect. We parameterise this alteration with integers $s$, where a positive number controls the number of components split, and negative numbers correspond to the number of components set to the background.
    \item The third kind of alteration is \emph{temperature}. Its purpose is to model how the entropy of the categorical distribution modelled by the segmentation network might affect the loss. For this, we increase the temperature $\tau$ in softmax operation $softmax(\sfrac{l}{\tau})$ for logits $l$ calculated from the input mask, which results in increasingly ``soft'' masks. 
\end{itemize}


The three types of alteration are composed to generate a synthetic prediction mask that can be used to investigate how the trajectory loss behaves as the mask changes. We use 25 trials to estimate the value of the loss for a given configuration of the parameters $s,\eta,\tau$.

\section{Implementation details}\label{sec:details}

Here, we further specify the configuration and implementation details used in our experiments.

\subsection{Extracting flow}

We use RAFT~\cite{teed2020raft}\footnote{Code and models available \url{https://github.com/princeton-vl/RAFT/tree/master} under BSD-3 license.} to extract optical flow pretrained on FlyingThings3D~\cite{mayer2016large}. We follow the methods used to extract flow in previous work~\cite{choudhury+karazija22gwm,yang2021self-supervised}. Namely, we consider pairs of frames with a distance in time of either 1 or 2, both in forward and backward directions for DAVIS and SegTrackv2. For FBMS, we consider distances of 3 and 6 due to lower motion setting in the dataset. The optical flow is extracted before training.

\subsection{Extracting trajectories}

We use CoTracker~\cite{karaev2023cotracker} to extract point trajectories. CoTracker is trained on MOVi-F Kubric~\cite{greff2022kubric} datasets. We use CoTracker v2\footnote{Code and models available \url{https://github.com/facebookresearch/co-tracker} under non-comercial license.}. We query at every 4th-pixel coordinate for each frame to extract point trajectories. At $480\times854$ resolution for DAVIS, this results in 25k points for each frame. When tracking, we find it beneficial to inject auxiliary query points. For this, we define two additional query grids with a stride of 32, querying a frame seven frames in the future and the past (or less if at the video boundaries). This generates around 2k additional points, which we do not use for training. When processing videos of heterogeneous resolutions, we resize the input to $480\times854$ to maintain the same number of points.

\subsection{Training hyperparameters}

For the segmentation network, we use the same model architecture as in \cite{choudhury+karazija22gwm} -- MaskFormer with DINO backbone.
We feed images at $192 \times 352$ resolution. We also use random horizontal flipping augmentation.
The network is trained to predict $k=4$ components, which, in the case of binary segmentation, are then merged into two following \cite{choudhury+karazija22gwm}.
We train using AdamW optimiser, with a learning rate of 1.5e-4, weight decay of 0.01, a batch size of 8, and a linear learning warmup schedule for 1500 iterations.
We train for 5000 iterations.\footnote{We estimate about 3 hours to train a model using A6000 GPU (peak GPU memory ~25GB). We estimate around 100 GPU hours to train models for the results here.}  We use an Exponential Moving Average (EMA) with the decay power of 2/3 with a warmup of 1500 iterations and update every 10 steps to help stabilise the training. On SegTrackv2 we instead used decay power 4/5 as the dataset is considerably smaller than others. We set $\lambda_f = 0.03$, $\lambda_t=5*10^-5$, and $\lambda_\tau=0.1$ in all experiments, which yields loss values in a similar numerical range. For the temporal smoothing loss, we use $\Delta t = 5$.

\subsection{MOVi-F experiments}

When conducting experiments on the MOVi-F dataset (Sec. 4.2), we consider ground-truth trajectories obtained from modified rendering script~\cite{greff2022kubric}.
We normalise the trajectories to the $[0, 1]$ range based on image width and height.

For K-Means, we consider the trajectories with the initial position at $T=0$ subtracted, thus clustering offsets from the initial position.

For SSC~\cite{elhamifar2013sparse}, we translate the method to Python following the original implementation in Matlab\footnote{Code available at \url{http://www.vision.jhu.edu/code/}}. We use the ADMM variant, which we found to give better results. We set the $\alpha=100$ and kept the rest of the hyperparameters unchanged. To transform the coefficient matrix into a graph adjacency, we found that simple symmetrisation yielded slightly better results than the proposed method that additionally normalised and filtered values. We report results for this method using the optimal number of clusters for spectral clustering.

For LRR~\cite{lu2012robust}, we similarly translate the method to Python following the original implementation in Matlab\footnote{Code available at \url{https://sites.google.com/site/guangcanliu/}}. We use $\lambda=0.2$. Additionally, we found it beneficial to reduce $\rho=1.01$ and use a larger number of iterations (10k) than proposed. Similarly to SSC, we experimented with different ways to transform the coefficient matrix to adjacency, including automatically determining the number of clusters based on the block-diagonal structure. We found, however, that using simpler symmetrisation with optimal numbers of clusters determined by an oracle gave the best results.

When considering our trajectory loss, we parameterise the masks with a small randomly initialised Unet~\cite{ronneberger2015u} predicting a 25-way segmentation, which we optimize using AdamW optimizer.

Note that K-Means, SSC and LRR baselines cluster trajectories rather than segmenting the image. To map back to the image domain and obtain segmentation masks, we repeatedly apply the method for each frame within a sequence, considering the trajectory for each pixel. This enables the most direct way to establish segmentation of the images through significant additional computation effort. An alternative could be to consider sequence wide-trajectories jointly; however, approaches like SSC and LLR do not scale well to such a large number of trajectories. For our trajectory loss, optimisation can be performed per sequence and, as we show in our real-world experiments, dataset-wide.

\newpage
\section*{NeurIPS Paper Checklist}

\begin{enumerate}

\item {\bf Claims}
    \item[] Question: Do the main claims made in the abstract and introduction accurately reflect the paper's contributions and scope?
    \item[] Answer: \answerYes{} %
    \item[] Justification: We have confirmed the viability of our loss formulation in controlled simulated settings, per-sequence optimisation settings with no tracking noise and in real-world settings. We also considered alternative formulations of the trajectories and found them to underperform. 

\item {\bf Limitations}
    \item[] Question: Does the paper discuss the limitations of the work performed by the authors?
    \item[] Answer: \answerYes{} %
    \item[] Justification: See \cref{sec:limitations}.

\item {\bf Theory Assumptions and Proofs}
    \item[] Question: For each theoretical result, does the paper provide the full set of assumptions and a complete (and correct) proof?
    \item[] Answer: \answerNA{} %
    \item[] Justification: the paper does not include theoretical results.

    \item {\bf Experimental Result Reproducibility}
    \item[] Question: Does the paper fully disclose all the information needed to reproduce the main experimental results of the paper to the extent that it affects the main claims and/or conclusions of the paper (regardless of whether the code and data are provided or not)?
    \item[] Answer: \answerYes{} %
    \item[] Justification: We list all relevant details in \cref{sec:details}.

\item {\bf Open access to data and code}
    \item[] Question: Does the paper provide open access to the data and code, with sufficient instructions to faithfully reproduce the main experimental results, as described in supplemental material?
    \item[] Answer: \answerNo{} %
    \item[] Justification: We do not include code at the time of submission but commit to releasing the code and models at a later time.

\item {\bf Experimental Setting/Details}
    \item[] Question: Does the paper specify all the training and test details (e.g., data splits, hyperparameters, how they were chosen, type of optimizer, etc.) necessary to understand the results?
    \item[] Answer: \answerYes{} %
    \item[] Justification: We include brief summary of key details in \cref{sec:experiments} and complete information in \cref{sec:details}.

\item {\bf Experiment Statistical Significance}
    \item[] Question: Does the paper report error bars suitably and correctly defined or other appropriate information about the statistical significance of the experiments?
    \item[] Answer: \answerNo{} %
    \item[] Justification: We do not include confidence intervals when reporting main experimental results due to the computational burden of doing so. We report $\pm\sigma$ intervals in our feasibility study.

\item {\bf Experiments Compute Resources}
    \item[] Question: For each experiment, does the paper provide sufficient information on the computer resources (type of compute workers, memory, time of execution) needed to reproduce the experiments?
    \item[] Answer: \answerYes{} %
    \item[] Justification: We give the computation cost of a single experiment and estimate total GPU hours required to train models for the results.
    
\item {\bf Code Of Ethics}
    \item[] Question: Does the research conducted in the paper conform, in every respect, with the NeurIPS Code of Ethics \url{https://neurips.cc/public/EthicsGuidelines}?
    \item[] Answer: \answerYes{} %
    \item[] Justification: We make use of publicly available and open-source code and models, respecting individual licenses.

\item {\bf Broader Impacts}
    \item[] Question: Does the paper discuss both potential positive societal impacts and negative societal impacts of the work performed?
    \item[] Answer: \answerYes{} %
    \item[] Justification: See \cref{sec:impact}.
    
\item {\bf Safeguards}
    \item[] Question: Does the paper describe safeguards that have been put in place for responsible release of data or models that have a high risk for misuse (e.g., pretrained language models, image generators, or scraped datasets)?
    \item[] Answer: \answerNA{} %
    \item[] Justification: As our key proposal is a method for learning segmentation using trajectory data, we do not foresee our models trained for benchmark datasets requiring safeguards as their use is limited due to the small dataset scale.

\item {\bf Licenses for existing assets}
    \item[] Question: Are the creators or original owners of assets (e.g., code, data, models), used in the paper, properly credited and are the license and terms of use explicitly mentioned and properly respected?
    \item[] Answer: \answerYes{} %
    \item[] Justification: We cite the models we build upon which are released on open licenses permitting such use.

\item {\bf New Assets}
    \item[] Question: Are new assets introduced in the paper well documented and is the documentation provided alongside the assets?
    \item[] Answer: \answerNA{} %
    \item[] Justification: The paper does not release new assets.

\item {\bf Crowdsourcing and Research with Human Subjects}
    \item[] Question: For crowdsourcing experiments and research with human subjects, does the paper include the full text of instructions given to participants and screenshots, if applicable, as well as details about compensation (if any)? 
    \item[] Answer: \answerNA{} %
    \item[] Justification: The paper does not involve crowdsourcing nor research with human subjects.

\item {\bf Institutional Review Board (IRB) Approvals or Equivalent for Research with Human Subjects}
    \item[] Question: Does the paper describe potential risks incurred by study participants, whether such risks were disclosed to the subjects, and whether Institutional Review Board (IRB) approvals (or an equivalent approval/review based on the requirements of your country or institution) were obtained?
    \item[] Answer: \answerNA{} %
    \item[] Justification: The paper does not involve crowdsourcing nor research with human subjects.

\end{enumerate}

\end{document}